\documentclass[11pt,a4paper]{article}
\usepackage[hyperref]{acl2019}
\usepackage{amsmath,amssymb}
\usepackage{times}

\usepackage{amsmath,amsfonts,bm}

\def\eqref#1{equation~\ref{#1}}
\def\1{\bm{1}}

\DeclareMathAlphabet{\mathsfit}{\encodingdefault}{\sfdefault}{m}{sl}
\SetMathAlphabet{\mathsfit}{bold}{\encodingdefault}{\sfdefault}{bx}{n}

\DeclareMathOperator*{\argmax}{arg\,max}

\usepackage{color,xcolor}
\usepackage{epsfig}
\usepackage{graphicx}

\usepackage{adjustbox}
\usepackage{array}
\usepackage{booktabs}
\usepackage{colortbl}
\usepackage{float,wrapfig}
\usepackage{hhline}
\usepackage{multirow}
\usepackage{amsmath,amsfonts,amssymb}
\usepackage{bm}
\usepackage{nicefrac}
\usepackage{microtype}

\usepackage{pifont}% http://ctan.org/pkg/pifont

\usepackage{changepage}
\usepackage{extramarks}
\usepackage{fancyhdr}
\usepackage{lastpage}
\usepackage{setspace}
\usepackage{soul}
\usepackage{xspace}

\usepackage{url}

\usepackage{enumerate}
\usepackage{todonotes} % conflict with CVPR/ICCV format

\usepackage{titlesec}

\usepackage{makecell}

\usepackage{pifont} % for extra symbols such as \cmark
\usepackage{subcaption}

\newcolumntype{L}[1]{>{\raggedright\let\newline\\\arraybackslash\hspace{0pt}}m{#1}}
\newcolumntype{C}[1]{>{\centering\let\newline\\\arraybackslash\hspace{0pt}}m{#1}}
\newcolumntype{R}[1]{>{\raggedleft\let\newline\\\arraybackslash\hspace{0pt}}m{#1}}

\newcommand{\sect}[1]{Section~\ref{#1}}

\newcommand{\eqn}[1]{Equation~\ref{#1}}

\newcommand{\fig}[1]{Figure~\ref{#1}}

\newcommand{\tbl}[1]{Table~\ref{#1}}

\newcommand{\ignorethis}[1]{}

\newcommand{\cmark}{\ding{51}}%
\newcommand{\xmark}{\ding{55}}

\makeatletter
\DeclareRobustCommand\onedot{\futurelet\@let@token\@onedot}
\def\@onedot{\ifx\@let@token.\else.\null\fi\xspace}

\def\eg{e.g\onedot} 
\def\ie{i.e\onedot}

\def\wrt{w.r.t\onedot} 

\makeatother

\definecolor{MyDarkBlue}{rgb}{0,0.08,1}
\definecolor{MyDarkGreen}{rgb}{0.02,0.6,0.02}
\definecolor{MyDarkRed}{rgb}{0.8,0.02,0.02}
\definecolor{MyDarkOrange}{rgb}{0.40,0.2,0.02}
\definecolor{MyPurple}{RGB}{111,0,255}
\definecolor{MyRed}{rgb}{1.0,0.0,0.0}
\definecolor{MyGold}{rgb}{0.75,0.6,0.12}
\definecolor{MyDarkgray}{rgb}{0.66, 0.66, 0.66}

\newcommand{\modelfull}{Visually Grounded Neural Syntax Learner\xspace}
\newcommand{\model}{VG-NSL\xspace}
\newcommand{\modelhi}{VG-NSL+HI\xspace}
\newcommand{\modelhift}{VG-NSL+HI+FastText\xspace}
\newcommand{\myparagraph}[1]{\vspace{-5pt}\paragraph{#1}}
\newcommand{\mycell}[2][c]{\begin{tabular}[#1]{@{}l@{}}#2\end{tabular}}

\usepackage{latexsym}
\usepackage{lipsum}
\usepackage{booktabs}
\usepackage{amsmath,amssymb}
\usepackage{times}
\usepackage{synttree}
\usepackage{parsetree}
\usepackage[counterclockwise]{rotating}
\usepackage[ruled,vlined,noresetcount]{algorithm2e}

\newif\ifdraft
\draftfalse
\definecolor{dkgreen}{RGB}{0,130,0}

\aclfinalcopy % Uncomment this line for the final submission
\def\aclpaperid{861} %  Enter the acl Paper ID here

\title{Visually Grounded Neural Syntax Acquisition}

\author{Haoyue Shi$^{\dagger,}$\thanks{~~HS and JM contributed equally to the work.} \qquad \quad
Jiayuan Mao$^{\ddagger,*}$ \qquad \quad
Kevin Gimpel$^{\dagger}$ \qquad \quad
Karen Livescu$^{\dagger}$ \\[0.02in]
$\dagger$: Toyota Technological Institute at Chicago, IL, USA \\
$\ddagger$: ITCS, Institute for Interdisciplinary Information Sciences, Tsinghua University, China \\
{\tt \{freda, kgimpel, klivescu\}@ttic.edu, mjy14@mails.tsinghua.edu.cn}  \\
}

\date{}

\begin{document}
\maketitle
\begin{abstract}
We present the \modelfull (\model), an approach for learning syntactic representations and structures without explicit supervision. The model learns by looking at natural images and reading paired captions. \model generates constituency parse trees of texts, recursively composes representations for constituents, and matches them with images. We define the {\it concreteness} of constituents by their matching scores with images, and use it to guide the parsing of text.
Experiments on the MSCOCO data set show that \model outperforms various unsupervised parsing approaches that do not use visual grounding, in terms of $\text{F}_1$ scores against gold parse trees. We find that \model is much more stable with respect to the choice of random initialization and the amount of training data.
We also find that the {\it concreteness} acquired by \model correlates well with a similar measure defined by linguists.
Finally, we also apply \model to multiple languages in the Multi30K data set, showing that our model consistently outperforms prior unsupervised approaches.\footnote{~Project page: \url{https://ttic.uchicago.edu/~freda/project/vgnsl}} 
\end{abstract}

\begin{figure}[t]
\centering
% \begin{subfigure}[t]{0.5\textwidth}
% \includegraphics[width=\textwidth]{fig/intro-NP.pdf}
% \caption{Contrastive learning for noun phrases. }
% \end{subfigure}
% \begin{subfigure}[t]{0.5\textwidth}
% \includegraphics[width=\textwidth]{fig/intro-PP.pdf}
% \caption{Contrastive learning for prepositional phrases. }
% \end{subfigure}
\includegraphics[width=0.40\textwidth]{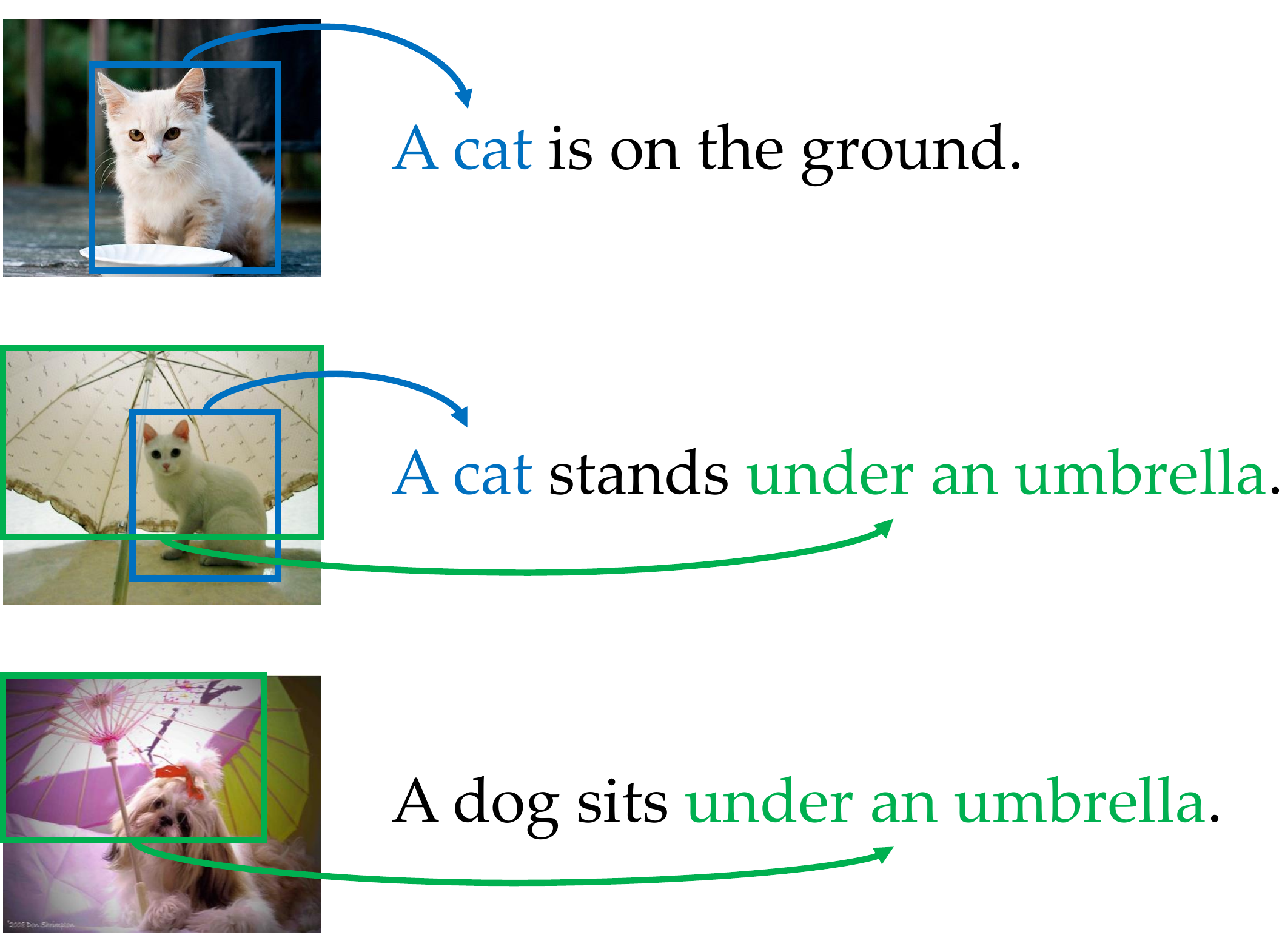}
\vspace{-0.5em}
\caption{\label{fig:intro} We propose to use image-caption pairs to extract constituents from text, based on the assumption that similar spans should be matched to similar visual objects and these concrete spans form constituents.}

\iffalse \caption{\label{fig:intro} We propose to use contrastive image-caption pairs to induce constituency parse trees. \model recursively combines consecutive constituents so that the combined constituents are contrastive with constituents matched with other images.}
\fi
\vspace{-1.5em}
\end{figure}

\section{Introduction}
We study the problem of visually grounded syntax acquisition.  Consider the images in \fig{fig:intro}, paired with the descriptive texts (captions) in English. Given no prior knowledge of English, and sufficient such pairs, one can infer the correspondence between certain words and visual attributes, (e.g., recognizing that ``\textit{a cat}'' refers to the objects in the blue boxes). One can further extract constituents, by assuming that concrete spans of words should be processed as a whole, and thus form the constituents. Similarly, the same process can be applied to verb or prepositional phrases.

\begin{figure*}
    \centering
    \includegraphics[width=0.9\textwidth]{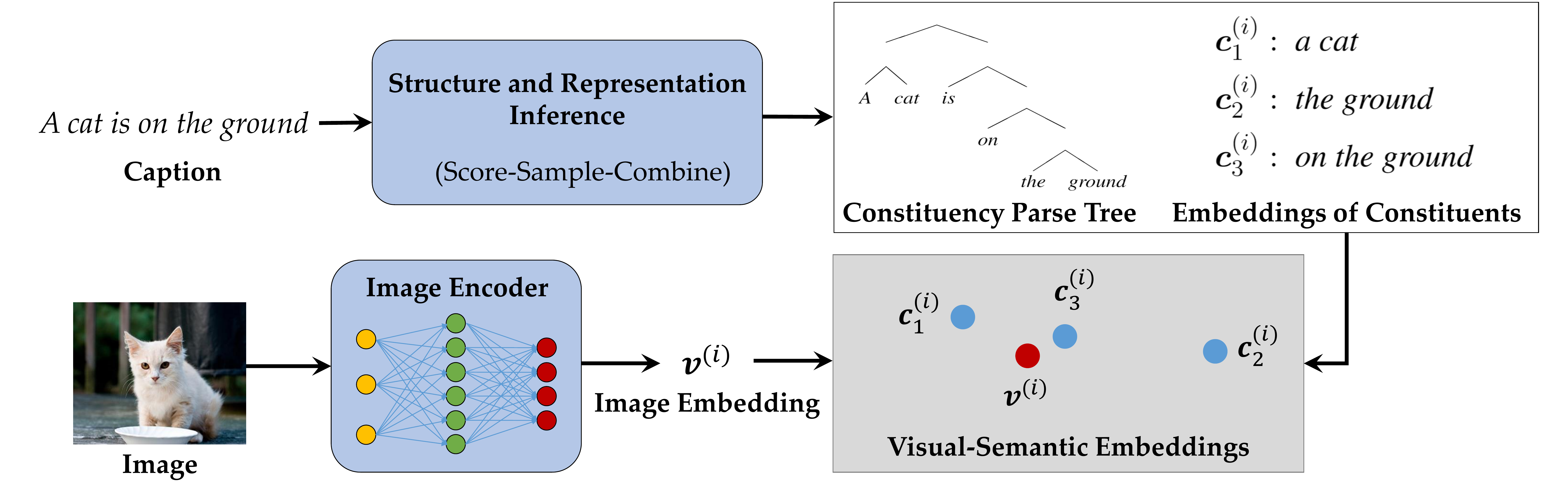}
    \vspace{-0.5em}
    \caption{\model consists of two modules: a textual module for inferring structures and representations for captions, and a visual-semantic module for matching constituents with images. \model induces constituency parse trees of captions by looking at images and reading paired captions.}
    \label{fig:model}
    \vspace{-1.2em}
\end{figure*}

This intuition motivates the use of image-text pairs to facilitate automated language learning, including both syntax and semantics. In this paper we focus on learning syntactic structures, and propose the \modelfull (\model, shown in \fig{fig:model}). \model acquires syntax, in the form of constituency parsing, by looking at images and reading captions.

At a high level, \model builds latent constituency trees of word sequences and recursively composes representations for constituents. Next, it matches the visual and textual representations.
The training procedure is built on the hypothesis that a better syntactic structure contributes to a better representation of constituents, which then leads to better alignment between vision and language.  We use no human-labeled constituency trees or other syntactic labeling (such as part-of-speech tags).
Instead, we define a {\it concreteness} score of constituents based on their matching with images, and use it to guide the parsing of sentences. At test time, no images paired with the text are needed. 

We compare \model with prior approaches to unsupervised language learning, most of which do not use visual grounding.
Our first finding is that \model improves over the best previous approaches to unsupervised constituency parsing in terms of $\text{F}_1$ scores against gold parse trees.
We also find that many existing approaches are quite unstable with respect to the choice of random initialization, whereas \model exhibits consistent parsing results across multiple training runs.
Third, we analyze the performance of different models on different types of constituents, and find that our model shows substantial improvement on noun phrases and prepositional phrases which are common in captions.
Fourth, \model is much more data-efficient than prior work based purely on text, achieving comparable performance to other approaches using only 20\% of the training captions.
In addition, the {\it concreteness} score, which emerges during the matching between constituents and images, correlates well with a similar measure defined by linguists.
Finally, \model can be easily extended to multiple languages, which we evaluate on the Multi30K data set \cite{elliott2016multi30k,elliott2017findings} consisting of German and French image captions.

\section{Related Work}
\paragraph{Linguistic structure induction from text.}
Recent work has proposed several approaches for inducing latent syntactic structures, including constituency trees  \citep{choi2018learning,yogatama2016learning,maillard2018latent,havrylov2019cooperative,kim2019unsupervised,drozdov2019unsupervised} and dependency trees \citep{shi2019learning}, from the distant supervision of downstream tasks. However, most of the methods are not able to produce linguistically sound structures, or even consistent ones with fixed data and hyperparameters but different random initializations  \citep{williams2018latent}. 

A related line of research is to induce latent syntactic structure via language modeling.  This approach has achieved remarkable performance on unsupervised constituency parsing \cite{shen2017neural, shen2019ordered}, especially in identifying the boundaries of higher-level (\ie, larger) constituents. 
To our knowledge, the Parsing-Reading-Predict Network \citep[PRPN;][]{shen2017neural} and the Ordered Neuron LSTM 
\citep[ON-LSTM;][]{shen2019ordered} currently produce the best fully unsupervised constituency parsing results.
One issue with PRPN, however, is that it tends to produce meaningless parses for lower-level (smaller) constituents \cite{htut2018grammar}. 

Over the last two decades, there has been extensive study targeting unsupervised constituency parsing \citep{klein2002generative,klein2004corpus,klein2005natural,bod2006all,bod2006unsupervised,ponvert2011simple} and dependency parsing \citep{klein2004corpus,smith-eisner:2006:COLACL,spitkovsky-alshawi-jurafsky:2010:NAACLHLT,han2017dependency}. However, all of these approaches are based on  linguistic annotations. Specifically, they operate on the part-of-speech tags of words instead of word tokens. 
One exception is \citet{spitkovsky2011unsupervised}, which produces dependency parse trees based on automatically induced pseudo tags. 

In contrast to these existing approaches, we focus on inducing constituency parse trees with visual grounding. We use parallel data from another modality (\ie, paired images and captions), instead of linguistic annotations such as POS tags. We include a detailed comparison between some related works in the supplementary material.

There has been some prior work on improving unsupervised parsing by leveraging extra signals, such as parallel text~\citep{snyder-naseem-barzilay:2009:ACLIJCNLP}, annotated data in another language with parallel text~\citep{ganchev-gillenwater-taskar:2009:ACLIJCNLP}, annotated data in other languages without parallel text~\citep{cohen-das-smith:2011:EMNLP}, or non-parallel text from multiple languages~\citep{cohen-smith:2009:NAACLHLT09}. We leave the integration of other grounding signals as future work.

\myparagraph{Grounded language acquisition.}
Grounded language acquisition has been studied for image-caption data \citep{christie-EtAl:2016:EMNLP2016}, video-caption data \citep{siddharth2014seeing,yu2015compositional}, and visual reasoning \citep{mao2019neurosymbolic}. However, existing approaches rely on human labels or rules for classifying visual attributes or actions. Instead, our model induces syntax structures with no human-defined labels or rules. 

Meanwhile, learning visual-semantic representations in a joint embedding space \citep{ngiam2011multimodal} is a widely studied approach, and has achieved remarkable results on image-caption retrieval \citep{kiros2014unifying,faghri2017vse++,shi2018learning}, image caption generation \citep{kiros2014unifying,karpathy2015deep,ma2015multimodal}, and visual question answering \citep{malinowski2015ask}.
In this work, we borrow this idea to match visual and textual representations.

\myparagraph{Concreteness estimation.} 
\citet{turney2011literal} define concrete words as those referring to things, events, and properties that we can perceive directly with our senses. 
Subsequent work has studied word-level concreteness estimation based on text \citep{turney2011literal,hill2013concreteness}, human judgments \citep{silberer2012grounded,hill2014concreteness,brysbaert2014concreteness}, and multi-modal data \citep{hill2014learning,hill2014multi,kiela2014improving,young2014image,hessel2018quantifying,silberer2017visually,bhaskar2017exploring}. 
As with \citet{hessel2018quantifying} and \citet{kiela2014improving}, our model uses multi-modal data to estimate concreteness.
Compared with them, we define concreteness for spans instead of words, and use it to induce linguistic structures. 

\section{\modelfull}
Given a set of paired images and captions, our goal is to learn  representations and structures for words and constituents.
Toward this goal, we propose the \modelfull (\model), an approach for the grounded acquisition of syntax of natural language.
\model is inspired by the idea of semantic bootstrapping \citep{pinker1984language}, which suggests that children acquire syntax by first understanding the meaning of words and phrases, and linking them with the syntax of words.

At a high level (Figure~\ref{fig:model}), \model consists of 2 modules. First, given an input caption (i.e., a sentence or a smaller constituent), as a sequence of tokens, \model builds a latent constituency parse tree, and recursively composes representations for every constituent. Next, it 
matches textual representations with visual inputs, such as the paired image with the constituents. Both modules are jointly optimized from natural supervision: the model acquires constituency structures, composes textual representations, and links them with visual scenes, by looking at images and reading paired captions.

\subsection{Textual Representations and Structures}
\model starts by composing a binary constituency structure of text, using an easy-first bottom-up parser \cite{goldberg2010efficient}. The composition of the tree from a caption of length $n$ consists of $n - 1$ steps. Let $\mathbf{X}^{(t)} = (\mathbf{x}^{(t)}_1, \mathbf{x}^{(t)}_2, \cdots, \mathbf{x}^{(t)}_k)$ denote the textual representations of a sequence of constituents after step $t$, where $k = n - t$.
For simplicity, we use $\mathbf{X}^{(0)}$ to denote the \textit{word embeddings} for all tokens (the initial representations).

At step $t$, a score function $\textit{score}(\cdot; \Theta)$, parameterized by $\Theta$, is evaluated on all pairs of consecutive constituents,
resulting in a vector $\textit{\textbf{score}}(\mathbf{X}^{(t-1)}; \Theta)$ of length $n - t$:
\begin{align*}
\textit{score}&(\mathbf{X}^{(t-1)};\Theta)_j\\
&\triangleq \textit{score}\left(\left[\mathbf{x}^{(t-1)}_j,\mathbf{x}^{(t-1)}_{j + 1}\right];\Theta\right).
\end{align*}
We implement $score(\cdot; \Theta)$ as a two-layer feed-forward network.

A pair of constituents $\left(\mathbf{x}^{(t-1)}_{j^*}, \mathbf{x}^{(t-1)}_{j^* + 1}\right)$ is sampled from all pairs of consecutive constituents, with respect to the distribution produced by a {\tt softmax}:\footnote{~At test time, we take the {\tt argmax}.}
\begin{align*}
    \Pr\left[j^*\right] &
    = \frac{\exp\left(\textit{score}\left(\mathbf{X}^{(t-1)}; \Theta\right)_{j^*}\right)}{
    \sum_{j} \exp\left(\textit{score}\left(\mathbf{X}^{(t-1)}; \Theta\right)_j\right)
    } .
\end{align*}
The selected pair is combined to form a single new constituent. Thus, after step $t$, the number of constituents is decreased by 1.  The textual representation for the new constituent is defined as the L2-normed sum of the two component constituents:
\begin{align*}
    \textit{combine}\left(\mathbf{x}^{(t-1)}_{j^*}, \mathbf{x}^{(t-1)}_{j^*+1}\right) \triangleq \frac{\mathbf{x}^{(t-1)}_{j^*} + \mathbf{x}^{(t-1)}_{j^*+1}}{\left\|\mathbf{x}^{(t-1)}_{j^*} + \mathbf{x}^{(t-1)}_{j^*+1}\right\|_2}.
\end{align*}
We find that using a more complex encoder for constituents, such as GRUs, will cause the representations to be highly biased towards a few salient words in the sentence \citep[\eg, the encoder encodes only the word ``cat'' while ignoring the rest part of the caption;][]{shi2018learning,wu2019unified}. This significantly degrades the performance of linguistic structure induction.

\begin{figure}[t]
    \centering
    \includegraphics[width=0.35\textwidth]{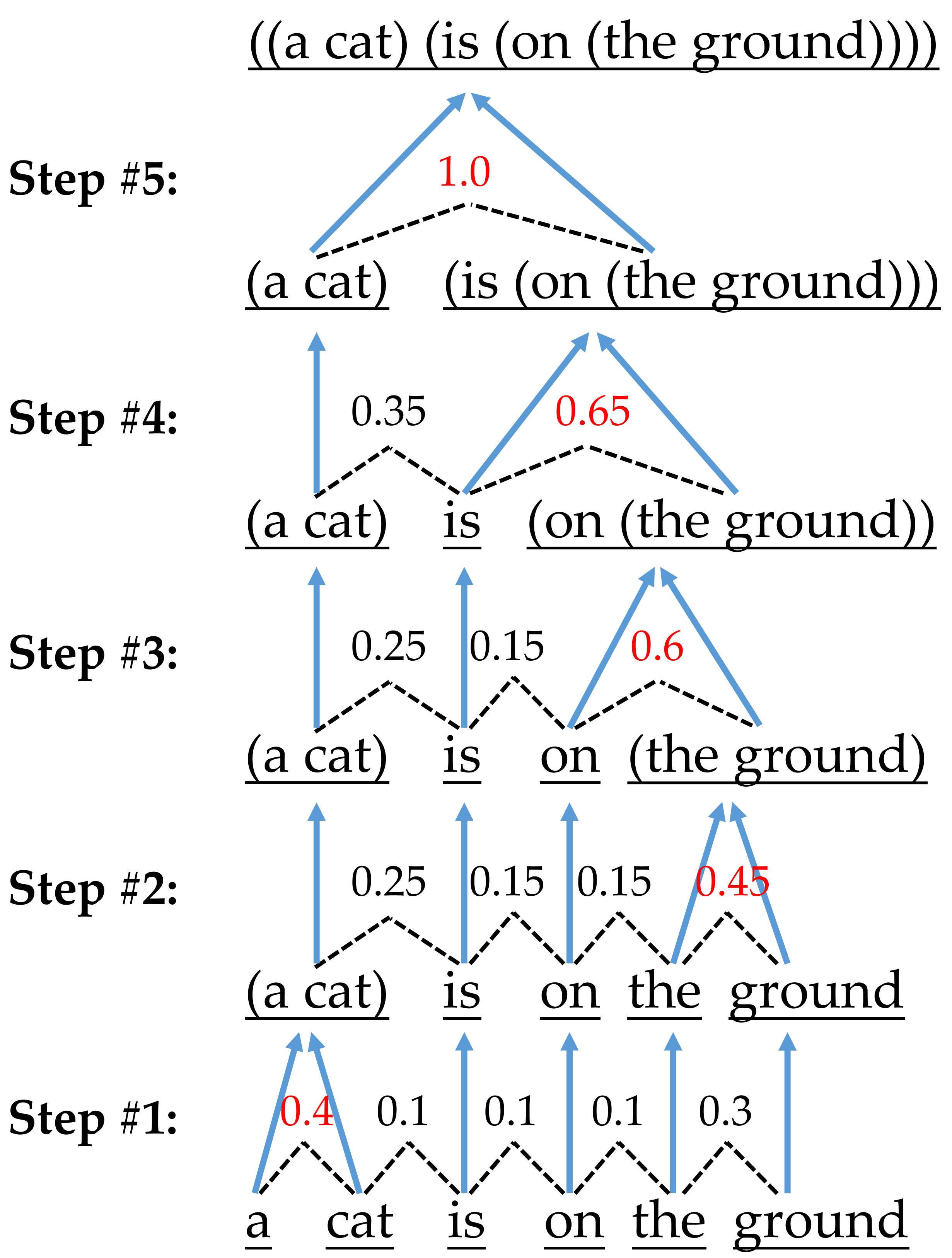}\vspace{-0.5em}
    \caption{An illustration of how \model composes a constituency parse tree. 
    At each step, the score function $score$ is evaluated on all pairs of consecutive constituents (dashed lines). Next, a pair of constituents is sampled from all pairs \wrt a distribution computed by the {\tt softmax} of all predicted scores. The selected pair of constituents is combined into a larger one, while the other constituents remain unchanged (solid lines).}\vspace{-1.5em} 
    
    % , by passing the concatenation of the feature vectors to the scorer and predicting a scalar score for each pair of adjacent constituents (dashed lines). 
    % Then the probability (numbers in the figure) of selecting a specific pair is computed by {\tt softmax} across the scores during training, where we take the {\tt argmax} when testing. 
    \label{figure:parsing-module}
\end{figure}

We repeat this score-sample-combine process for $n - 1$ steps, until all words in the input text have been combined into a single constituent (Figure~\ref{figure:parsing-module}). This ends the inference of the constituency parse tree.
Since at each time step we combine two consecutive constituents, the derived tree $\mathbf{t}$ contains $2n - 1$ constituents (including all words).

\subsection{Visual-Semantic Embeddings}
We follow an approach similar to that of \citet{kiros2014unifying} to define the visual-semantic embedding (VSE) space for paired images and text constituents. 
Let $\mathbf{v}^{(i)}$ denote the vector representation of an image $i$, and $\mathbf{c}^{(i)}_{j}$ denote the vector representation of the $j$-th constituent of its corresponding text caption. During the matching with images, we ignore the tree structure and index them as a list of constituents.
A function $m(\cdot, \cdot; \Phi)$ is defined as the matching score between images and texts:
\begin{align*}
    m(\mathbf{v}^{(i)}, \mathbf{c}^{(i)}_j; \Phi) &\triangleq \cos(\Phi \mathbf{v}, \mathbf{c}),
\end{align*}
where the parameter vector $\Phi$ aligns the visual and textual representations into a joint space.

\subsection{Training}
We optimize the visual-semantic representations ($\Phi$) and constituency structures ($\Theta$) in an alternating approach.
At each iteration, given constituency parsing results of caption, $\Phi$ is optimized for matching the visual and the textual representations. Next, given the visual grounding of constituents, $\Theta$ is optimized for producing constituents that can be better matched with images. Specifically, we optimize textual representations and the visual-semantic embedding space using a hinge-based triplet ranking loss:
\begin{align}
\begin{aligned}
    & \mathcal{L}(\Phi; \mathcal{V}, \mathcal{C}) = \\
    & \sum_{\begin{subarray}{c}
    i, k \neq i,
    j, \ell
\end{subarray}} \left[m(\mathbf{c}^{(k)}_{\ell}, \mathbf{v}^{(i)}) - m(\mathbf{c}^{(i)}_{j}, \mathbf{v}^{(i)}) + \delta\right]_+\\
    + & \sum_{\begin{subarray}{c}
    i, k \neq i,
    j
\end{subarray}}\left[m(\mathbf{c}^{(i)}_{j}, \mathbf{v}^{(k)}) - m(\mathbf{c}^{(i)}_{j}, \mathbf{v}^{(i)}) + \delta\right]_+,
\end{aligned}
\nonumber
\end{align}
where $i$ and $k$ index over all image-caption pairs in the data set, while $j$ and $\ell$ enumerate all constituents of a specific caption ($c^{(i)}$ and $c^{(k)}$, respectively), $\mathcal{V} = \{\mathbf{v}^{(i)}\}$ is the set of image representations, $\mathcal{C} = \{\mathbf{c}^{(i)}_{j}\}$ is the set of textual representations of all constituents, and $\delta$ is a constant margin, $[\cdot]_+$ denotes $\max(0, \cdot)$. The loss $\mathcal{L}$ extends the loss for image-caption retrieval of \citet{kiros2014unifying}, by introducing the alignments between images and sub-sentence constituents.

We optimize textual structures via distant supervision: they are optimized for a better alignment between the derived constituents and the images. Intuitively, the following objective encourages adjectives to be associated (combined) with the corresponding nouns, and verbs/prepositions to be associated (combined) with the corresponding subjects and objects.
Specifically, we use REINFORCE \citep{williams1992simple} as the gradient estimator for $\Theta$.
Consider the parsing process of a specific caption $c^{(i)}$, and denote the corresponding image embedding $\mathbf{v}^{(i)}$. For a constituent $\mathbf{z}$ of $c^{(i)}$, we define its (visual) concreteness $\textit{concrete}(\mathbf{z}, \mathbf{v}^{(i)})$ as: 
\begin{align}
    & \textit{concrete}(\mathbf{z}, \mathbf{v}^{(i)}) = \nonumber \\
    & \sum_{k\neq i, p} \left[m(\mathbf{z}, \mathbf{v}^{(i)}) - m(\mathbf{c}^{(k)}_{p}, \mathbf{v}^{(i)}) - \delta'\right]_+ \nonumber \\
    + & \sum_{k\neq i} \left[m(\mathbf{z}, \mathbf{v}^{(i)}) - m(\mathbf{z}, \mathbf{v}^{(k)}) - \delta'\right]_+,
    \label{eq:concreteness}
\end{align}
where $\delta'$ is a fixed margin. At step $t$, we define the reward function for a combination of a pair of constituents ($\mathbf{x}^{(t-1)}_j$, $\mathbf{x}^{(t-1)}_{j + 1}$) as:
\begin{align}
    & r(\mathbf{x}^{(t-1)}_j, \mathbf{x}^{(t-1)}_{j + 1}) = \textit{concrete}(\mathbf{z}, \mathbf{v}^{(i)}),
\label{eq:reward}
\end{align} 
where $\mathbf{z} \triangleq \textit{combine}(\mathbf{x}^{(t-1)}_j, \mathbf{x}^{(t-1)}_{j + 1})$. In plain words, at each step, we encourage the model to compose a constituent that maximizes the alignment between the new constituent and the corresponding image. During training, we sample constituency parse trees of captions, and reinforce each composition step using \eqn{eq:reward}. During test, no paired images of text are needed.

\subsection{The Head-Initial Inductive Bias}
English and many other Indo-European languages are usually head-initial \cite{baker2001atoms}. For example, in verb phrases or prepositional phrases, the verb (or the preposition) precedes the complements (\eg, the object of the verb). Consider the simple caption {\it a white cat on the lawn}. While the association of the adjective ({\it white}) could be induced from the visual grounding of phrases, whether the preposition ({\it on}) should be associated with {\it a white cat} or {\it the lawn} is more challenging to induce. 
Thus, we impose an inductive bias to guide the learner to correctly associate prepositions with their complements, determiners with corresponding noun phrases, and 
complementizers with the corresponding relative clauses. Specifically, we discourage 
abstract constituents (\ie, constituents that cannot be grounded in the image) from being combined with a preceding constituent, by modifying the original reward definition (\eqn{eq:reward}) as:
\begin{equation}
    \begin{aligned}
    r'(\mathbf{x}^{(t-1)}_j , & \mathbf{x}^{(t-1)}_{j + 1}) \\
    = & \frac{r(\mathbf{x}^{(t-1)}_j, \mathbf{x}^{(t-1)}_{j + 1})}{\lambda \cdot \textit{abstract}(\mathbf{x}^{(t-1)}_{j + 1}, \mathbf{v}^{(i)}) + 1} \ ,
\end{aligned}
\label{eq:inductive-bias}
\end{equation}
where $\lambda$ is a scalar hyperparameter, $\mathbf{v}^{(i)}$ is the image embedding corresponding to the caption being parsed, and $abstract$ denotes the {\it abstractness} of the span, defined analogously to concreteness (\eqn{eq:concreteness}):
\begin{align*}
    & \textit{abstract}(\mathbf{z}, \mathbf{v}^{(i)}) = \nonumber \\
    & \sum_{k\neq i, p} \left[m(\mathbf{c}^{(k)}_{p}, \mathbf{v}^{(i)})  -m(\mathbf{z}, \mathbf{v}^{(i)}) + \delta'\right]_+\nonumber \\
    + & \sum_{k\neq i} \left[m(\mathbf{z}, \mathbf{v}^{(k)}) -m(\mathbf{z}, \mathbf{v}^{(i)}) + \delta'\right]_+,
\end{align*}

The intuition here is that the initial heads for prepositional phrases (e.g., {\it on}) and relative clauses (e.g., {\it which, where}) are usually abstract words. During training, we encourage the model to associate these abstract words with the succeeding constituents instead of the preceding ones.
It is worth noting that such an inductive bias is language-specific, and cannot be applied to head-final languages such as Japanese \cite{baker2001atoms}. We leave the design of head-directionality inductive biases for other languages as future work.

\section{Experiments}
We evaluate \model for unsupervised parsing in a few ways: $\text{F}_1$ score with gold trees, self-consistency across different choices of random initialization, performance on different types of constituents, and data efficiency. In addition, we find that the {\it concreteness} score acquired by \model is  consistent with a similar measure defined by linguists.  We focus on English for the main experiments, but also extend to German and French.

\subsection{Data Sets and Metrics}
We use the standard split of the MSCOCO data set \cite{lin2014microsoft},
following \newcite{karpathy2015deep}.
It contains 82,783 images for training, 1,000 for development, and another 1,000 for testing. 
Each image is associated with 5 captions. 

For the evaluation of constituency parsing, the Penn Treebank \cite[PTB;][]{marcus1993building} is a widely used, manually annotated data set.
However, PTB consists of sentences from abstract domains, \eg, the {\it Wall Street Journal} (WSJ), which are not visually grounded and whose linguistic structures can hardly be induced by \model. 
Here we evaluate models on the MSCOCO test set, which is well-matched to the training domain; we leave the extension of our work to more abstract domains to future work. 
We apply Benepar \cite{kitaev2018constituency},\footnote{~\url{https://pypi.org/project/benepar}} an off-the-shelf constituency parser with state-of-the-art performance (95.52 $\text{F}_1$ score) on the WSJ test set,\footnote{~We also manually label the constituency parse trees for 50 captions randomly sampled from the MSCOCO test split, where Benepar has an $\text{F}_1$ score of 95.65 with the manual labels. Details can be found in the supplementary material. } to parse the captions in the MSCOCO test set as gold constituency parse trees.
We evaluate all of the investigated models using the $\text{F}_1$ score compared to these gold parse trees.\footnote{~Following convention \cite{sekine1997evalb}, we report the $\text{F}_1$ score across all constituents in the data set, instead of the average of sentence-level $\text{F}_1$ scores.}

\subsection{Baselines}
\label{sec:baseline}
\begin{table*}[t]
\centering
\setlength{\tabcolsep}{0.5pt}
\begin{tabular}{lrlrlrlrlrlr}
    \toprule
    \textbf{Model} & \multicolumn{2}{c}{\textbf{NP}} & \multicolumn{2}{c}{\textbf{VP}} & \multicolumn{2}{c}{\textbf{PP}} & \multicolumn{2}{c}{\textbf{ADJP}} & \multicolumn{2}{c}{\textbf{Avg. $\textbf{F}_1$}} & \multicolumn{1}{c}{\textbf{Self $\textbf{F}_1$}} \\
    \midrule
    Random  & 47.3&$_{\pm 0.3}$ & 10.5&$_{\pm 0.4}$ & 17.3&$_{\pm 0.7}$ & 33.5&$_{\pm 0.8}$ & 27.1&$_{\pm 0.2}$ & 32.4 \\
    Left    & 51.4&& 1.8 && 0.2 && 16.0 && $23.3$ && N/A\\
    Right   & 32.2&& 23.4&& 18.7 && 14.4 && $22.9$ && N/A\\
    PMI     & 54.2&& 16.0 && 14.3 && 39.2 && $30.5$ && N/A\\
    PRPN \citep{shen2017neural}    & ~~~72.8&$_{\pm 9.7}$~~~ & ~~~33.0&$_{\pm 9.1}$ ~~~& ~~~61.6&$_{\pm 9.9}$ ~~~& ~~~35.4&$_{\pm 4.3}$ ~~~& ~~~52.5&$_{\pm 2.6}$ ~~~& 60.3 \\
    ON-LSTM \citep{shen2019ordered}     & 74.4&$_{\pm 7.1}$ & 11.8&$_{\pm 5.6}$ & 41.3&$_{\pm 16.4}$ & \textbf{44.0}&$_{\pm 14.0}$ &  45.5&$_{\pm 3.3}$ & 69.3 \\
    Gumbel \citep{choi2018learning}$^\dagger$ & 50.4&$_{\pm 0.3}$ & 8.7&$_{\pm 0.3}$ & 15.5&$_{\pm 0.0}$ & 34.8&$_{\pm 1.6}$ & 27.9&$_{\pm 0.2}$ & 40.1 \\
    \midrule
    \model (ours)$^\dagger$ & \textbf{79.6}&$_{\pm 0.4}$ & 26.2&$_{\pm 0.4}$ & 42.0&$_{\pm 0.6}$ & 22.0&$_{\pm 0.4}$ & 50.4&$_{\pm 0.3}$ & 87.1\\
    \modelhi (ours)$^\dagger$ & 74.6&$_{\pm 0.5}$ & 32.5&$_{\pm 1.5}$ & \textbf{66.5}&$_{\pm 1.2}$ & 21.7&$_{\pm 1.1}$ & \textbf{53.3}&$_{\pm 0.2}$ & \textbf{90.2} \\
    \modelhift (ours)*$^\dagger$ & 78.8&$_{\pm 0.5}$ & 24.4&$_{\pm 0.9}$ & 65.6&$_{\pm 1.1}$ & 22.0&$_{\pm 0.7}$ & \textbf{54.4}&$_{\pm 0.4}$ & 89.8 \\
    \toprule
    \multicolumn{12}{c}{\textit{ Concreteness estimation--based models}} \\
    \midrule
    \citet{turney2011literal}* &  65.5 && 30.8 && 35.3 && 30.4 && 42.5 && N/A\\
    \citet{turney2011literal}+HI*  & 74.5 && 26.2  && 47.6 && 25.6 && 48.9 && N/A \\
    \citet{brysbaert2014concreteness}* & 54.1 && 27.8 && 27.0 && 33.1 &&	34.1 && N/A \\
    \citet{brysbaert2014concreteness}+HI* & 73.4 && 23.9 && 50.0 && 26.1 && 47.9 && N/A \\
    \citet{hessel2018quantifying}$^\dagger$ & 50.9 && 21.7 &&	32.8 && 27.5 && 33.2  && N/A \\
    \citet{hessel2018quantifying}+HI$^\dagger$ &	72.5 && \textbf{34.4} &&	65.8 && 26.2 &&	52.9 && N/A \\
    % problematic results
    % \citet{turney2011literal}* &  55.7 && 5.7 && 14.4 && 22.1 && 29.1 && N/A\\
    % \citet{turney2011literal}+HI*  & \textbf{77.4} && \textbf{38.5} && 62.8 && 25.9 && \textbf{54.4} && N/A \\
    % \citet{brysbaert2014concreteness}* & 52.4	&& 25.6 &&	29.0 && 33.6 &&	33.1 && N/A \\
    % \citet{brysbaert2014concreteness}+HI* & 72.0 &&22.8 && 38.7 && 26.5 && 44.9 && N/A \\
    % \citet{hessel2018quantifying} &     50.4 && 21.0 &&	 32.4 && 27.2 && 32.7  && N/A \\
    % \citet{hessel2018quantifying}+HI & 72.3	&& 34.4 &&	65.8 && 25.8 &&	52.8 && N/A \\
    % Semi-supervised* &  55.7 && 5.7 && 14.4 && 22.1 && 29.1 && N/A\\
    % Semi-supervised+HI*  & \textbf{77.4} && \textbf{38.5} && 62.8 && 25.9 && \textbf{54.4} && N/A \\
    % Manually-labeled & 52.4	&& 25.6 &&	29.0 && 33.6 &&	33.1 && N/A \\
    % Manually-labeled+HI* & 72.0 &&22.8 && 38.7 && 26.5 && 44.9 && N/A \\
    \bottomrule
\end{tabular}\vspace{-0.4em}
\caption{\label{table:main-result}
% \karen{will results will pre-trained word embeddings still be added?}\freda{Yes, coming soon, as the previously trained models were using the same random seeds by mistake. We are re-training it.} \freda{I deleted the dumped random trees by mistake, updating it with the newly generated ones -- should not affect the analysis and conclusions at all.}
Recall of specific typed phrases, and overall $\text{F}_1$ score, evaluated on the MSCOCO test split, averaged over 5 runs with different random initializations. We also include self-agreement $\text{F}_1$ score \cite{williams2018latent} across the 5 runs. 
$\pm$ denotes standard deviation.
* denotes models requiring extra labels and/or corpus, and $\dagger$ denotes models requiring a pre-trained visual feature extractor.
We highlight the best number in each column among all models that do not require extra data other than paired image-caption data, as well as the overall best number.
The Left, Right, PMI, and concreteness estimation--based models have no standard deviation or self $\text{F}_1$ (shown as N/A) as they are deterministic given the training and/or testing data.
% \freda{done, updated.} \freda{I was writing the algorithm in supplementary material for concreteness score based approaches, and realized that the concreteness to unseen words I previously gave is not consistent across different concreteness scores. I gave unk words a concreteness of 0, but the range by \newcite{turney2011literal} is (-1, 1), the range of \newcite{brysbaert2014concreteness} is (1,5) and that of \newcite{hessel2018quantifying} is (0, inf). I cannot reproduce such a high score by \newcite{turney2011literal} with either associate unk word with mean(scores) or min(scores). Sorry about that, and we might probably want to rewrite some paragraphs analyzing this, but I think we should give a stable algorithm for all the concreteness baselines. I'll update the result soon, with assuming unks are the most abstract words -- this is consistent with our model, where unk cannot be matched to anything. } 
% \karen{would be good to align number columns so the decimal points are aligned vertically.}\freda{resolved}
% \freda{added another baseline based on fully manually-labeled concreteness, not sure if we could refer to the models by citet, as they are originally not doing grammar induction (any other suggestion?).}
}
\vspace{-1.2em}
\end{table*}

We compare \model with various baselines for unsupervised tree structure modeling of texts. We can categorize the baselines by their training objective or supervision.

\paragraph{Trivial tree structures.} 
Similarly to recent work on latent tree structures \cite{williams2018latent,htut2018grammar,shi2018tree}, we include three types of {\it trivial} baselines without linguistic information: random binary trees, left-branching binary trees, and right-branching binary trees. 

\paragraph{Syntax acquisition by language modeling and statistics.} \citet{shen2017neural} proposes the Parsing-Reading-Predict Network (PRPN), which predicts syntactic distances \citep{shen2018straight} between adjacent words, and composes a binary tree based on the syntactic distances to improve language modeling. The learned distances can be mapped into a binary constituency parse tree, by recursively splitting the sentence between the two consecutive words with the largest syntactic distance.

Ordered neurons \citep[ON-LSTM;][]{shen2019ordered}
is a recurrent unit based on the LSTM cell \citep{hochreiter1997long}  that explicitly regularizes different neurons in a cell to represent short-term or long-term information.
After being trained on the language modeling task, \citet{shen2019ordered} suggest that the gate values in ON-LSTM cells can be viewed as syntactic distances \cite{shen2018straight} between adjacent words to %further 
induce latent tree structures. 
ON-LSTM has the state-of-the-art unsupervised constituency parsing performance on the WSJ test set. We train both PRPN and ON-LSTM on all captions in the MSCOCO training set and use the models as baselines.

Inspired by the syntactic distance--based approaches \citep{shen2017neural,shen2019ordered}, we also introduce another baseline, PMI, which uses negative pointwise mutual information \cite{ChurchHanks90} between adjacent words as the syntactic distance. We compose constituency parse trees based on the distances in the same way as PRPN and ON-LSTM.

\paragraph{Syntax acquisition from downstream tasks.}
\newcite{choi2018learning} propose to compose binary constituency parse trees directly from downstream tasks using the Gumbel softmax trick \cite{jang2016categorical}.
We integrate a Gumbel tree-based caption encoder into the visual semantic embedding approach \cite{kiros2014unifying}. The model is trained on the downstream task of image-caption retrieval.

\paragraph{Syntax acquisition from concreteness estimation.}
Since we apply concreteness information 
to train \model, it is worth comparing against unsupervised constituency parsing based on previous approaches for predicting word concreteness. This set of baselines includes semi-supervised estimation \cite{turney2011literal}, crowdsourced labeling \cite{brysbaert2014concreteness}, and multimodal estimation \cite{hessel2018quantifying}. Note that none of these approaches has been applied to unsupervised constituency parsing.
Implementation details can be found in the supplementary material.  

Based on the concreteness score of words, we introduce another baseline 
similar to \model. Specifically, we recursively combine two consecutive constituents with the largest average concreteness, and use the average concreteness as the score for the composed constituent.
The algorithm generates binary constituency parse trees of captions.
For a fair comparison, we implement a variant of this algorithm that also uses a head-initial inductive bias and include the details in the appendix.

\subsection{Implementation Details}
Across all experiments and all models (including baselines such as PRPN, ON-LSTM, and Gumbel), the embedding dimension for words and constituents is 512. 
For \model, we use a pre-trained ResNet-101 \citep{he2016deep}, trained on ImageNet \citep{russakovsky2015imagenet}, to extract vector embeddings for images. Thus, $\Phi$ is a mapping from a 2048-D image embedding space to a 512-D visual-semantic embedding space. 
As for the $\textit{score}$ function in constituency parsing, we use a hidden dimension of 128 and ReLU activation.
All \model models are trained for 30 epochs. We use an Adam optimizer \citep{kingma2015adam} with initial learning rate $5\times 10^{-4}$ to train \model. 
The learning rate is re-initialized to $5 \times 10^{-5}$ after 15 epochs.
We tune other hyperparameters of \model on the development set using the self-agreement $\text{F}_1$ score \cite{williams2018latent} over 5 runs with different choices of random initialization.

\subsection{Results:  Unsupervised Constituency Parsing}

We evaluate the induced constituency parse trees via the overall $\text{F}_1$ score, as well as the recall of four types of constituents: noun phrases (NP), verb phrases (VP), prepositional phrases (PP), and adjective phrases (ADJP) (\tbl{table:main-result}). 
We also evaluate the robustness of models trained with fixed data and hyperparameters, but different random initialization, in two ways:  via the standard deviation of performance across multiple runs, and via the self-agreement $\text{F}_1$ score \cite{williams2018latent}, which is the average $\text{F}_1$ taken over pairs of different runs. 

Among all of the models which do not require extra labels, \model with the head-initial inductive bias (\modelhi) achieves the best $\text{F}_1$ score.
PRPN \cite{shen2017neural} and a concreteness estimation-based baseline \citep{hessel2018quantifying} both produce competitive results.
It is worth noting that the PRPN baseline reaches this performance without any information from images.
However, the performance of PRPN is less stable than that of \model across random initializations. 
In contrast to its state-of-the-art performance on the WSJ full set \citep{shen2019ordered}, we observe that ON-LSTM does not perform well on the MSCOCO caption data set. However, it remains the best model for adjective phrases, which is consistent with the result reported by \citet{shen2019ordered}.

In addition to the best overall $\text{F}_1$ scores, \modelhi achieves competitive scores across most phrase types (NP, VP and PP). Our models (\model and \modelhi) perform the best on
NP and PP, which are the most common visually grounded phrases in the MSCOCO data set.
In addition, our models produce much higher self $\text{F}_1$ than the baselines \cite{shen2017neural,shen2019ordered,choi2018learning}, showing that they reliably produce reasonable constituency parse trees with different initialization.

We also test the effectiveness of using pre-trained word embeddings. Specifically, for  \modelhift, we use a pre-trained FastText embedding (300-D, \citealp{joulin2016fasttext}), concatenated with a 212-D trainable embedding, as the word embedding. Using pre-trained word embeddings further improves performance to an average $\text{F}_1$ of 54.4\% while keeping a comparable self $\text{F}_1$.

\subsection{Results: Data Efficiency}
\label{sec:data-efficiency}
\begin{figure}[t]
\centering
\begin{subfigure}[t]{0.38\textwidth}
\centering
\includegraphics[width=\textwidth]{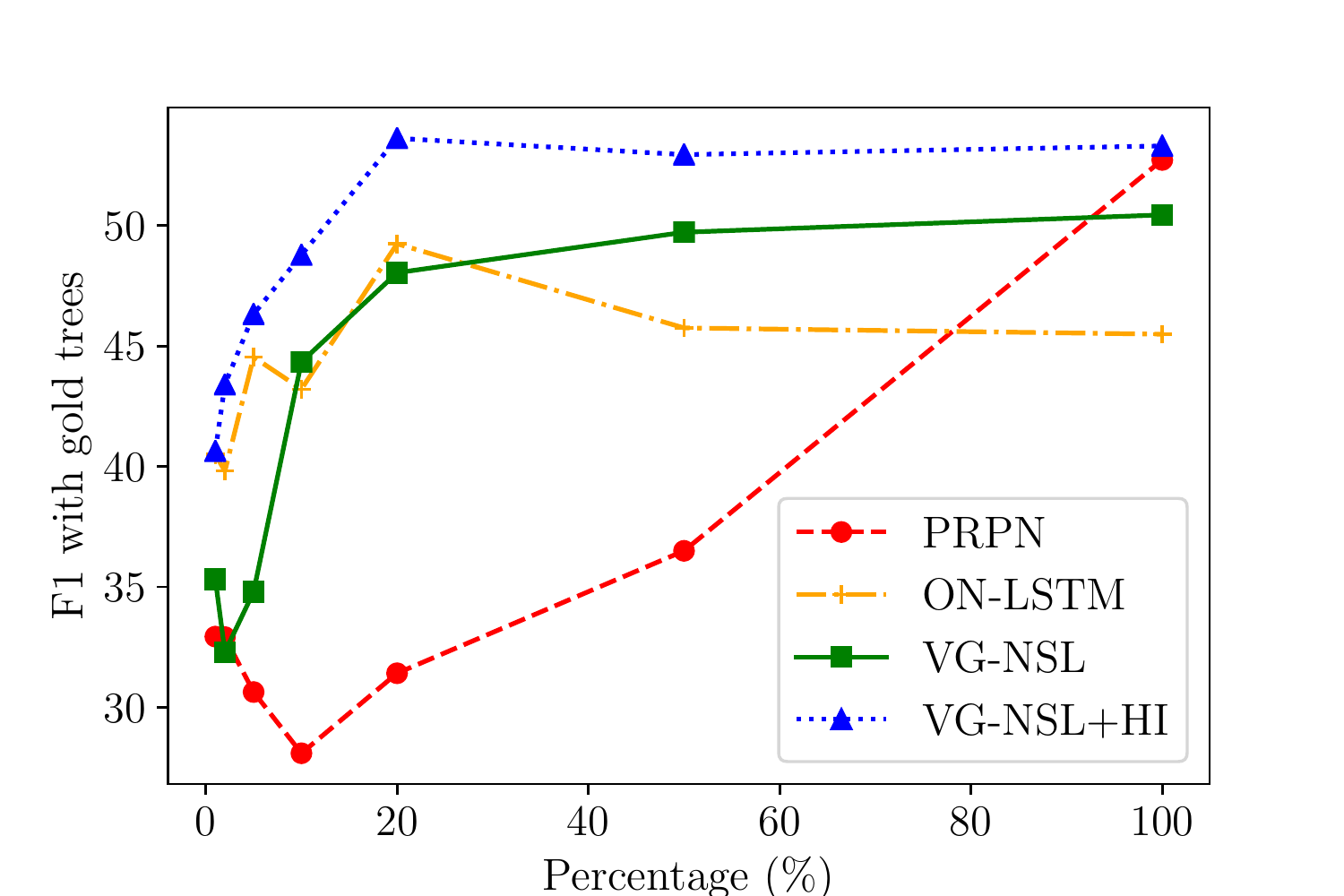}\vspace{-0.3em}
\caption{The percent data-$\text{F}_1$ curves.}\vspace{-0.2em}
\end{subfigure}
\begin{subfigure}[t]{0.38\textwidth}
\centering
\includegraphics[width=\textwidth]{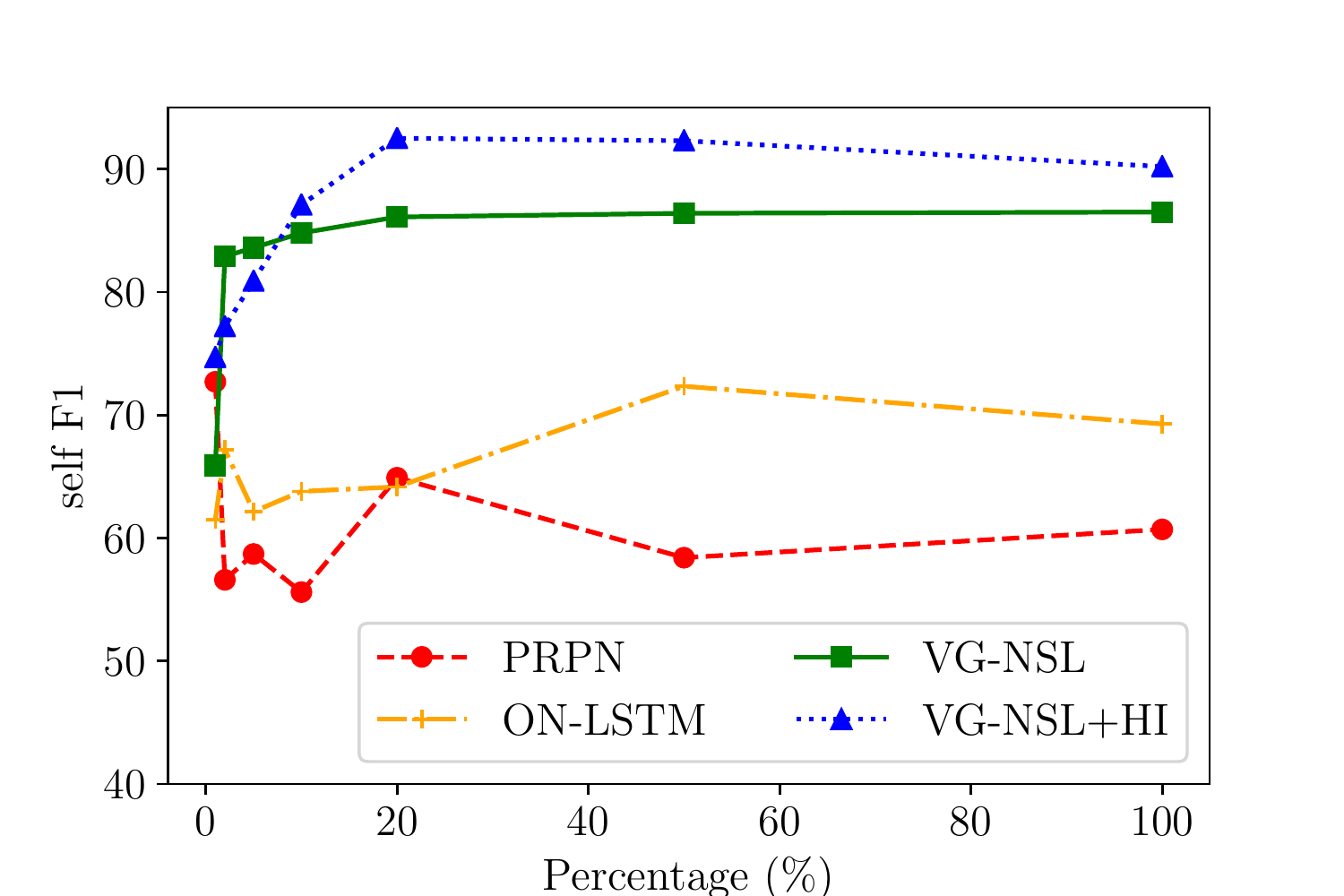}\vspace{-0.3em}
\caption{The percent data-self $\text{F}_1$ curves.}
\end{subfigure}
\vspace{-0.5em}
\caption{\label{fig:data-efficiency} $\text{F}_1$ score and self $\text{F}_1$ score with respect to the amount of training data. All numbers are averaged over 5 runs with different random initialization.
%The x axis denotes the amount of training data as a percentage of the full training set, while the y axis denotes the $\text{F}_1$ scores (left) and self $\text{F}_1$ scores (right).   %\freda{maybe smaller figures if space is limited.} \karen{yes that would work.  But axis labels should be larger font.} \freda{done.}
%\karen{when updating this fig, would be good to make it a bit larger (just reducing whitespace around them would help)}
}
\vspace{-1.5em}
\end{figure}

We compare the data efficiency for PRPN (the strongest baseline method), ON-LSTM,
\model, and \modelhi.
We train the models using 1\%, 2\%, 5\%, 10\%, 20\%, 50\% and 100\% of the MSCOCO training set, and report the overall $\text{F}_1$ and self $\text{F}_1$ scores on the test set (\fig{fig:data-efficiency}).

Compared to PRPN trained on the full training set, \model and \modelhi reach comparable performance using only 20\% of the data (\ie, 8K images with 40K captions).
\model tends to quickly become more stable (in terms of the self $\text{F}_1$ score) as the amount of data increases, while PRPN and ON-LSTM 
remain less stable.

\subsection{Analysis: Consistency with Linguistic Concreteness}
\begin{table}[t]
    \centering 
    \begin{tabular}{lcc}
    \toprule
        \textbf{Model/method} & \model & (+HI) \\
    \midrule
        \citet{turney2011literal} & 0.74 & 0.72 \\
        \citet{brysbaert2014concreteness} & 0.71 & 0.71 \\
        \citet{hessel2018quantifying} & 0.84 & 0.85 \\
    \bottomrule
    \end{tabular}
    \vspace{-0.5em}
    \caption{Agreement between our concreteness estimates and existing models or labels, evaluated via the Pearson correlation coefficient computed over the most frequent 100 words in the MSCOCO test set, averaged over 5 runs with different random initialization.}
    \label{tab:concreteness}
    \vspace{-0.5em}
\end{table}

During training, \model acquires concreteness estimates for constituents via \eqn{eq:concreteness}. Here, we evaluate the consistency between word-level concreteness estimates induced by \model and those produced by other methods \cite{turney2011literal,brysbaert2014concreteness,hessel2018quantifying}.
Specifically, we measure the correlation between the concreteness estimated by \model on MSCOCO test set and existing linguistic concreteness definitions (\tbl{tab:concreteness}). 
For any word, of which the representation is $\mathbf{z}$, we estimate its concreteness by taking the average of $\textit{concrete}(\mathbf{z}, \mathbf{v}^{(i)})$, across all associated images $\mathbf{v}^{(i)}$.
The high correlation between \model and the concreteness scores produced by \newcite{turney2011literal} and \newcite{brysbaert2014concreteness} supports the argument that the linguistic concept of concreteness can be acquired in an unsupervised way. Our model also achieves a high correlation with \citet{hessel2018quantifying}, which also estimates word concreteness based on visual-domain information. 

\subsection{Analysis: Self-Agreement $\text{F}_1$ Score as the Criterion for Model Selection}
\begin{table}[t]
    \centering
    \begin{tabular}{lccr}
    \toprule
    \bf Model & \bf Criterion & \bf Avg. $\textbf{F}_1$ & \bf Self $\textbf{F}_1$ \\
    \midrule
    \model & Self $\text{F}_1$ & \textbf{50.4} $_{\pm0.3}$ & \textbf{87.1} \\
    \model & R@1 & 47.7 $_{\pm0.6}$ & 83.4 \\
    \midrule
    \modelhi & Self $\text{F}_1$ & \textbf{53.3} $_{\pm0.2}$ & \textbf{90.2} \\
    \modelhi& R@1& 53.1 $_{\pm0.2}$ & 88.7 \\
    \bottomrule
    \end{tabular}
    \vspace{-0.5em}
    \caption{Average $\text{F}_1$ scores and Self $\text{F}_1$ scores of \model and \modelhi with different model selection methods. R@1 denotes using recall at 1 \cite{kiros2014unifying} as the model selection criterion. All hyperparameters are tuned with respect to self-agreement $\text{F}_1$ score. The numbers are comparable to those in Table~\ref{table:main-result}.} 
    \label{tab:self-agreement-F1}
    \vspace{-1.2em}
\end{table}
We introduce a novel hyperparameter tuning and model selection method based on the self-agreement $\text{F}_1$ score. 

Let $\mathcal{M}_\mathcal{H}^{(i, j)}$ denote the j-th checkpoint of the i-th model trained with hyperparameters $\mathcal{H}$, where $\mathcal{M}_\mathcal{H}^{(i_1, \cdot)}$ and  $\mathcal{M}_\mathcal{H}^{(i_2, \cdot)}$ differ in their random initialization. The hyperparameters $\mathcal{H}$ are tuned to maximize:
\begin{align*}
    \sum_{1\leq i < k \leq N} & \max_{|j_i - j_k| < \delta} \textit{F}_1\left(\mathcal{M}_\mathcal{H}^{(i, j_i)}, \mathcal{M}_\mathcal{H}^{(k, j_k)}\right),
\end{align*}
where $F_1(\cdot, \cdot)$ denotes the $\text{F}_1$ score between the trees generated by two models, $N$ the number of different runs, and $\delta$ the margin to ensure only nearby checkpoints are compared.\footnote{~In all of our experiments, $N=5, \delta=2$.}

After finding the best hyperparameters $\mathcal{H}_0$, we train the model for another $N$ times with different random initialization, and select the best models by 
\begin{align*}
    \argmax_{\{j_\ell \}_{\ell=1}^N} \sum_{1\leq i < k \leq N} \textit{F}_1\left(\mathcal{M}_{\mathcal{H}_0}^{(i, j_i)}, \mathcal{M}_{\mathcal{H}_0}^{(k, j_k)}\right).
\end{align*}

We compare the performance of \model selected by the self $\text{F}_1$ score and that selected by recall at 1 in image-to-text retrieval \citep[R@1 in \tbl{tab:self-agreement-F1};][]{kiros2014unifying}. As a model selection criterion, self $\text{F}_1$ consistently outperforms R@1 (avg. $\text{F}_1$: 50.4 {\it vs.} 47.7 and 53.3 {\it vs.} 53.1 for \model and \modelhi, respectively).
Meanwhile, it is worth noting that even if we select \model by R@1, it shows better stability compared with PRPN and ON-LSTM (Table~\ref{table:main-result}), in terms of the score variance across different random initialization and self $\text{F}_1$. Specifically, the variance of avg. $\text{F}_1$ is always less than 0.6 while the self $\text{F}_1$ is greater than 80.

Note that the PRPN and ON-LSTM models are not tuned using self $\text{F}_1$, since these models are usually trained for hundreds or thousands of epochs and thus it is computationally expensive to evaluate self $\text{F}_1$. We leave the efficient tuning of these baselines by self $\text{F}_1$ as a future work.

\subsection{Extension to Multiple Languages}
\begin{table}[t]
    \centering
    \setlength{\tabcolsep}{1pt}
    \begin{tabular}{lrlrlrl}
         \toprule
         \textbf{Model} & \multicolumn{2}{c}{\textbf{EN}} & \multicolumn{2}{c}{\textbf{DE}} & \multicolumn{2}{c}{\textbf{FR}} \\
         \midrule
          PRPN & ~~30.8&$_{\pm 17.9}$~ & ~~31.5&$_{\pm 8.9}$~ & ~~27.5&$_{\pm 7.0}$~ \\
          ON-LSTM   & ~~\textbf{38.7}&$_{\pm 12.7}$  &  ~~34.9&$_{\pm 12.3}$ & ~~27.7&$_{\pm 5.6}$~ \\ 
          \model & 33.5&$_{\pm 0.2}$ & 36.3&$_{\pm 0.2}$ & 34.3&$_{\pm 0.6}$ \\
          \modelhi & \textbf{38.7}&$_{\pm 0.2}$ & \textbf{38.3}&$_{\pm 0.2}$ & \textbf{38.1}&$_{\pm 0.6}$ \\
         \bottomrule
    \end{tabular}
    \vspace{-0.5em}
    \caption{$\text{F}_1$ scores on the Multi30K test split \cite{young2014image,elliott2016multi30k,elliott2017findings}, averaged over 5 runs with different random initialization. $\pm$ denotes the standard deviation.
    }
    \label{table:other-languages}
    \vspace{-1.5em}
\end{table}
We extend our experiments to the Multi30K data set, which is built on the Flickr30K data set \cite{young2014image} and consists of English, German \cite{elliott2016multi30k}, and French \cite{elliott2017findings} captions.
For Multi30K, there are 29,000 images in the training set, 1,014 in the  development set and 1,000 in the test set. Each image is associated with one caption in each language. 

We compare our models to PRPN and ON-LSTM in terms of overall $\text{F}_1$ score (\tbl{table:other-languages}).  \model with the head-initial inductive bias consistently performs the best across the three languages, all of which are highly head-initial \cite{baker2001atoms}. Note that the $\text{F}_1$ scores here are not comparable to those in \tbl{table:main-result}, since Multi30K (English) has 13x fewer captions than MSCOCO.

\section{Discussion}
We have proposed a simple but effective model, the \modelfull, for visually grounded language structure acquisition. 
\model jointly learns parse trees and visually grounded textual representations.  
In our experiments, we find that this approach to grounded language learning produces parsing models that are both accurate and stable, and that the learning is much more data-efficient than a state-of-the-art text-only approach.  Along the way, the model acquires estimates of word concreteness.

The results suggest multiple future research directions.  First, \model 
matches text embeddings directly with embeddings of entire images. 
Its performance may be boosted by considering structured representations of both images (\eg, \citealp{lu2016visual,wu2019unified}) and texts~\citep{steedman2000syntactic}.
Second, thus far we have used a shared representation for both syntax and semantics, but it may be useful to disentangle 
their representations~\citep{steedman2000syntactic}.
Third, our best model is based on the head-initial inductive bias.  Automatically acquiring such inductive biases from data remains challenging \citep{kemp2006learning,gauthier2018word}.
Finally, it may be possible to extend our approach to other linguistic tasks such as dependency parsing \citep{christie2016resolving}, coreference resolution \citep{kottur2018visual}, and learning pragmatics beyond semantics \citep{andreas2016reasoning}. 

There are also limitations to the idea of grounded language acquisition. In particular, the current approach has thus far been applied to understanding grounded texts in a single domain (static visual scenes for \model). Its applicability could be extended by learning shared representations across multiple modalities \citep{castrejon2016learning} or integrating with pure text-domain models (such as PRPN, \citealp{shen2017neural}).

\section*{Acknowledgement}
We thank Allyson Ettinger for helpful suggestions on this work, and the anonymous reviewers for their valuable feedback.

\bibliography{acl2019}
\bibliographystyle{acl_natbib}

~\\
\textbf{\Large Supplementary Material} \\
\appendix
The supplementary material is organized as follows. First, in \sect{appendix: metadata}, we summarize and compare existing models for constituency parsing without explicit syntactic supervision. Next, in \sect{appendix: implementation}, we present more implementation details of \model. Third, in \sect{appendix: baselines}, we present the implementation details for all of our baseline models. Fourth, in \sect{appendix:evaluation}, we present the evaluation details of Benepar \cite{kitaev2018constituency} on the MSCOCO data set. Fifth, in \sect{appendix: concreteness}, we qualitatively and quantitatively compare the concreteness scores estimated or labeled by different methods. Finally, in \sect{appendix: example-trees}, we show sample trees generated by VG-NSL on the MSCOCO test set.

\section{Overview of Models for Constituency Parsing without Explicit Syntactic Supervision}
\label{appendix: metadata}
\newcommand{\MAP}{MAP}
\begin{table*}[t]
\centering \small
\begin{tabular}{lllccc}
    \toprule \small
    \textbf{Model} & \textbf{Objective} & \textbf{Extra Label} & \textbf{Multi-} & \textbf{Stochastic} & \textbf{Extra}\\
    &&&\textbf{modal} & &\textbf{Corpus} \\
    \midrule
    CCM \cite{klein2002generative}* & \MAP & POS & \xmark &  \cmark & \xmark \\
    DMV-CCM \cite{klein2005natural}* & \MAP & POS & \xmark & \cmark  & \xmark \\
    U-DOP \cite{bod2006unsupervised}* & Probability Estimation & POS & \xmark & \xmark & \xmark \\
    UML-DOP \cite{bod2006all}* & \MAP & POS & \xmark & \cmark & \xmark \\
     \midrule
    PMI     & N/A & \xmark & \xmark & \xmark & \xmark \\
    Random  & N/A & \xmark & \xmark & \cmark & \xmark  \\
    Left    & N/A & \xmark & \xmark & \xmark & \xmark  \\
    Right   & N/A & \xmark & \xmark & \xmark & \xmark \\
    PRPN \cite{shen2017neural} & LM & \xmark & \xmark & \cmark & \xmark \\
    ON-LSTM \cite{shen2019ordered} & LM & \xmark & \xmark & \cmark & \xmark \\
    Gumbel softmax\cite{choi2018learning} & Cross-modal Retrieval & \xmark & \cmark & \cmark & \xmark \\
    \midrule
    \model (ours)  & Cross-modal Retrieval & \xmark & \cmark & \cmark & \xmark \\
    \modelhi (ours)  & Cross-modal Retrieval & \xmark & \cmark & \cmark & \xmark  \\
    \midrule
    \multicolumn{6}{c}{\textit{Concreteness estimation based models}} \\
    \midrule
    \citet{turney2011literal}*   & N/A & \mycell[t]{Concreteness\\(Partial)} & \xmark & \xmark& \cmark\\
    \citet{turney2011literal}+HI*   & N/A & \mycell[t]{Concreteness\\(Partial)} & \xmark & \xmark & \cmark\\
    \citet{brysbaert2014concreteness}*   & N/A & \mycell[t]{Concreteness\\(Full)} & \xmark & \xmark& \xmark \\
    \citet{brysbaert2014concreteness}+HI*   & N/A & \mycell[t]{Concreteness\\(Full)} & \xmark & \xmark & \xmark \\
    \citet{hessel2018quantifying}   & N/A & \xmark & \cmark & \xmark& \xmark \\
    \citet{hessel2018quantifying}+HI   & N/A & \xmark & \cmark & \xmark & \xmark \\
    \bottomrule
\end{tabular}
\caption{\label{table:model-summary} Comparison of models for constituency parsing without explicit syntactic supervision. * denotes models requiring extra labels, such as POS tags or manually labeled concreteness scores. All multimodal methods listed in the table require a pretrained visual feature extractor (i.e., ResNet-101; \citeauthor{he2016deep}, \citeyear{he2016deep}). A model is labeled as stochastic if for fixed training data and hyperparameters the model may produce different results (\eg, due to different choices of random initialization). To the best of our knowledge, results on concreteness estimation \cite{turney2011literal,brysbaert2014concreteness,hessel2018quantifying} have not been applied to unsupervised parsing so far.}
\end{table*}
\begin{figure*}[!t]
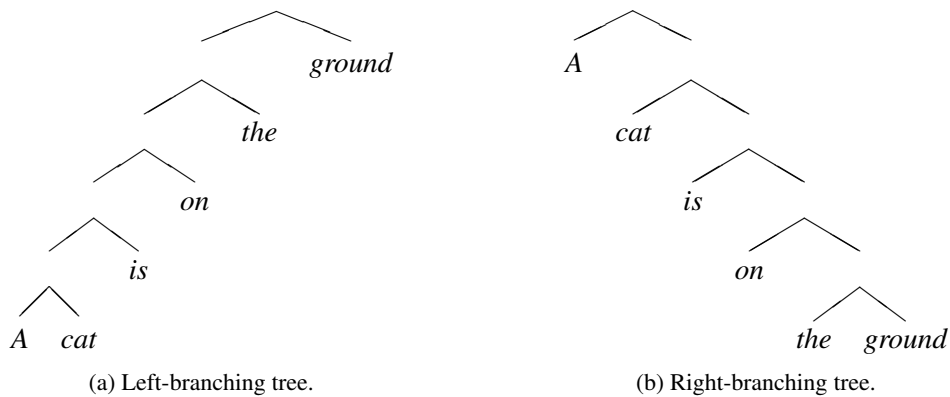

    \centering
    \begin{subfigure}[h]{0.45 \textwidth}
        \centering
        \begin{parsetree}
            ( .. 
                (.. 
                    (..
                        (..
                            (.. `A' `cat' ) 
                            `is' 
                        ) `on' 
                    ) `the' 
                ) `ground' 
            )
            
        \end{parsetree}
        \caption{Left-branching tree.}
    \end{subfigure}
    \begin{subfigure}[h]{0.45 \textwidth}
        \centering
        \begin{parsetree}
            (.. `A' (.. `cat'(.. `is' (.. `on'  ( .. `the'  `ground' ) ) ) ) )
        \end{parsetree}
        \caption{Right-branching tree.}
    \end{subfigure}
    
    \caption{Examples of some trivial tree structures.}
    \label{figure:trivial-trees}
\end{figure*}
Shown in \tbl{table:model-summary}, we compare existing models for constituency parsing without explicit syntactic supervision, with respect to their learning objective, dependence on extra labels or extra corpus, and other features. The table also includes the analysis of previous works on parsing sentences based on gold part-of-speech tags.

\section{Implementation Details for \model}
\label{appendix: implementation}

We adopt the code released by \citet{faghri2017vse++}\footnote{\url{https://github.com/fartashf/vsepp}} as the visual-semantic embedding module for \model. Following them, we fix the margin $\delta$ to $0.2$.
We also use the vocabulary provided by \citet{faghri2017vse++},\footnote{\url{http://www.cs.toronto.edu/~faghri/vsepp/vocab.tar}} which contains 10,000 frequent words in the MSCOCO data set. Out-of-vocabulary words are treated as unseen words. For either \model or baselines, we use the same vocabulary if applicable. 

\paragraph{Hyperparameter tuning.}
As stated in main text, we use the self-agreement $\text{F}_1$ score \cite{williams2018latent} as an unsupervised signal for tuning all hyperparamters. Besides the learning rate and other conventional hyperparameters, we also tune $\lambda$, the hyperparameter for the head-initial bias model. $\lambda$ indicates the weight of penalization for ``right abstract constituents''. We choose $\lambda$ from $\{1, 2, 5, 10, 20, 50, 100\}$ and found that $\lambda = 20$ gives the best self-agreement $\text{F}_1$ score.

\section{Implementation Details for Baselines}
\label{appendix: baselines}
\paragraph{Trivial tree structures.}
We show examples for left-branching binary trees and right-branching binary trees in \fig{figure:trivial-trees}. As for binary random trees, we iteratively combine two randomly selected adjacent constituents. This procedure is similar to that shown in Algorithm~\ref{algo:concreteness}.

\paragraph{Parsing-Reading-Predict Network (PRPN).}
We use the code released by \citet{shen2017neural} to train PRPN.\footnote{\url{https://github.com/yikangshen/PRPN}} We tune the hyperparameters with respect to language modeling perplexity \cite{jelinek1977perplexity}. For a fair comparison, we fix the hidden dimension of all hidden layers of PRPN as 512. We use an Adam optimizer \cite{kingma2015adam} to optimize the parameters. The tuned parameters are number of layers (1, 2, 3) and learning rate ($1\times 10^{-3}$, $5\times 10^{-4}$, $2 \times 10^{-4}$). The models are trained for 100 epochs on the MSCOCO dataset and 1,000 epochs on the Multi30K dataset, and are early stopped using the criterion of language model perplexity. 

\paragraph{Ordered Neurons (ON-LSTM).}
We use the code release by \citet{shen2019ordered} to train ON-LSTM.\footnote{\url{https://github.com/yikangshen/Ordered-Neurons}}
We tune the hyperparameters with respect to language modeling perplexity \cite{jelinek1977perplexity}, and use perplexity as an early stopping criterion. For a fair comparison, the hidden dimension of all hidden layers is set to 512, and the chunk size is changed to 16 to fit the hidden layer size.
Following the original paper \citep{shen2019ordered}, we set the number of layers to be 3, and report the constituency parse tree with respect to the gate values output by the second layer of ON-LSTM.
In order to obtain a better perplexity, we explore both Adam \cite{kingma2015adam} and SGD as the optimizer.
We tune the learning rate ($1\times 10^{-3}$, $5\times 10^{-4}$, $2 \times 10^{-4}$ for Adam, and $0.1$, $1$, $10$, $30$ for SGD).
The models are trained for 100 epochs on the MSCOCO dataset and 1,000 epochs on the Multi30K dataset, and are early stopped using the criterion of language model perplexity.

\paragraph{PMI based constituency parsing.}
We estimate the pointwise mutual information \citep[PMI;][]{ChurchHanks90} between two words using all captions in MSCOCO training set. We apply negative PMI as syntactic distance \cite{shen2018straight} to generate a binary constituency parse tree recursively.
The method of constituency parsing with a given list of syntactic distances is shown in Algorithm~\ref{algo:syntactic-distance}.
\begin{algorithm}[t]
\SetAlgoLined
\SetKwInOut{Input}{Output}
\SetKwFunction{parse}{parse}
\KwIn{text length $m$, list of syntactic distances $\boldsymbol{d} = (d_1, d_2, \ldots, d_{m-1})$}
\KwOut{Boundaries of constituents $B = \{(L_i, R_i)\}_{i=1, \ldots, m-1}$} 
$B$ = \parse{$\boldsymbol{d}$, $1$, $m$} \\~\\
\textbf{Function} \parse{$\boldsymbol{d}$, $\textit{left}$, $\textit{right}$} { \\
    \If{$\textit{left} = \textit{right}$}{
        \KwRet \FuncSty{EmptySet}
    }
    $p = \argmax_{j \in [\textit{left}, \textit{right-1}]} d_j$ \\
    $\textit{boundaries} = $ \FuncSty{union}(\\
    ~~ \{($\textit{left}$, $\textit{right}$)\},\\
    ~~\parse($\boldsymbol{d}$, \textit{left}, $p$), \\
    ~~\parse($\boldsymbol{d}$, $p+1$, \textit{right}) \\
    ) \\
    
    \KwRet \textit{boundaries}
}

 \caption{\label{algo:syntactic-distance} Constituency parsing based on given syntactic distance. }
\end{algorithm}

\paragraph{Gumbel-softmax based latent tree.}
We integrate Gumbel-softmax latent tree based text encoder \cite{choi2018learning}\footnote{\url{https://github.com/jihunchoi/unsupervised-treelstm}} to the visual semantic embedding framework \cite{faghri2017vse++}, and use the tree structure produced by it as a baseline. 

\begin{table*}[t]
    \centering \small
    \begin{tabular}{lrrrr}
    \toprule
        & \citet{turney2011literal} & \citet{brysbaert2014concreteness} & \citet{hessel2018quantifying} & \modelhi \\
    \midrule
         \citet{turney2011literal} &  1.00 & 0.84 & 0.58 & 0.72 \\
         \citet{brysbaert2014concreteness} &  0.84 & 1.00 & 0.55 & 0.71\\
         \citet{hessel2018quantifying} & 0.58 & 0.55 & 1.00 & 0.85\\
         \modelhi & 0.72 & 0.71 & 0.85 & 1.00 \\
    \bottomrule
    \end{tabular}
    \caption{Pearson correlation coefficients between existing concreteness estimation methods, including baselines and \modelhi. In order to make a fair comparison, the correlation coefficients are evaluated on the 100 most frequent words on MSCOCO test set.}
    \label{table:concreteness-others}
\end{table*}
\begin{figure*}[t]
    \centering
    \includegraphics[width=\textwidth]{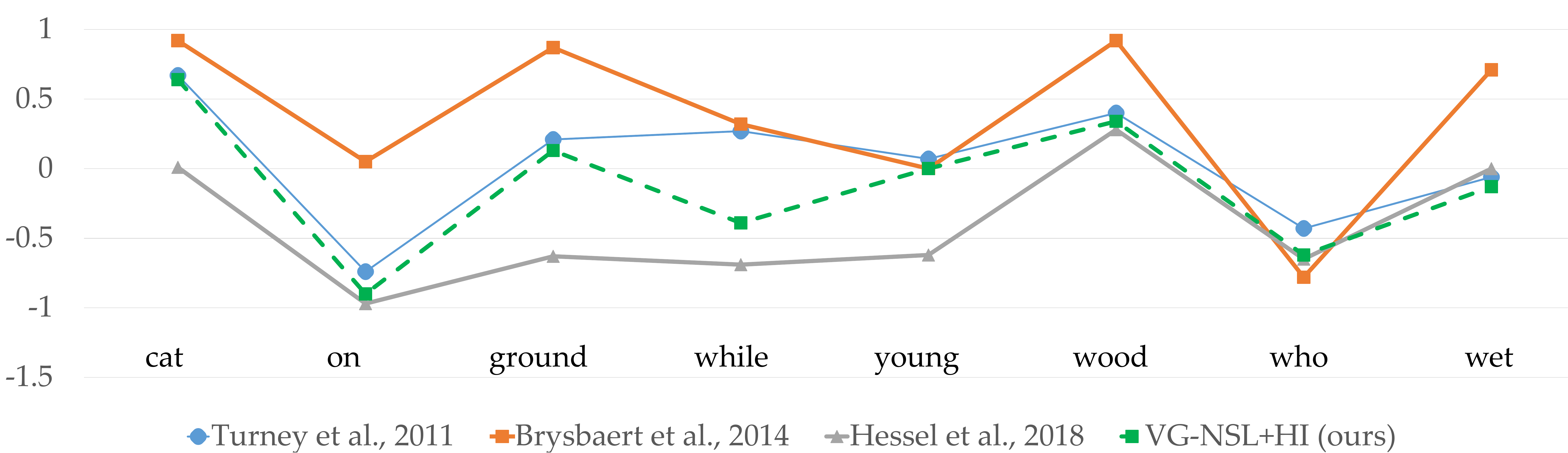}
    \caption{Normalized concreteness scores of example words.} 
    \label{figure:concreteness-words}
\end{figure*}
\paragraph{Concreteness estimation.}
For the semi-supervised concreteness estimation, we reproduce the experiments by \citet{turney2011literal}, applying the manually labeled concreteness scores for 4,295 words from the MRC Psycholinguistic Database Machine
Usable Dictionary \cite{coltheart1981mrc} as supervision,\footnote{\url{http://ota.oucs.ox.ac.uk/headers/1054.xml}} and use English Wikipedia pages to estimate PMI between words.\footnote{{\url{https://dumps.wikimedia.org/other/static_html_dumps/April_2007/en/}}} The PMI is then used to compute similarity between seen and unseen words, which is further used as weights to estimate concreteness for unseen words.
%\karen{say how PMI is used}\freda{resolved.}
For the concreteness scores from crowdsourcing, we use the released data set of \citet{brysbaert2014concreteness}.\footnote{\url{http://crr.ugent.be/archives/1330}}
Similarly to \model, the multimodal concreteness score \cite{hessel2018quantifying} is also estimated on the MSCOCO training set, using an open-sourced implementation.\footnote{\url{https://github.com/victorssilva/concreteness}}

\paragraph{Constituency parsing with concreteness scores.}
Denote $\alpha(w)$ as the concreteness score estimated by a model for the word $w$. Given a sequence of concreteness scores of caption tokens denoted by $(\alpha(w_1), \alpha(w_2), \ldots, \alpha(w_m))$, we aim to produce a binary constituency parse tree.
We first normalize the concreteness scores to the range of $[-1, 1]$, via:\footnote{~For the concreteness scores estimated by \citet{hessel2018quantifying}, we let $\alpha(w) = \log \alpha(w)$ before normalizing, as the original scores are in the range of $(0, +\infty)$.}
\begin{align}
    \alpha'(w_i) = \frac{2\left(\alpha(w_i) - \frac{\max_{j} \alpha(w_j) - \min_{j} \alpha(w_j)}{2}\right)}{\max_{j} \alpha(w_j) - \min_{j} \alpha(w_j)} ~.
    \label{eq:concreteness-normalization}
    \nonumber
\end{align}
We treat unseen words (i.e., out-of-vocabulary words) in the same way in \model, by assigning the concreteness of $-1$ to unseen words, with the assumption that unseen words are the most abstract ones. 

We compose constituency parse trees using the normalized concreteness scores by iteratively combining consecutive constituents. At each step, we select two adjacent constituents (initially, words) with the highest average concreteness score and combine them into a larger constituent, of which the concreteness is the average of its children. We repeat the above procedure until there is only one constituent left. 

As for the head-initial inductive bias, we weight the concreteness of the right constituent with a hyperparemeter $\tau > 1$ when ranking all pairs of consecutive constituents during selection. Meanwhile, the concreteness of the composed constituent remains the average of the two component constituents. In order to keep consistent with \model, we set $\tau=20$ in all of our experiments.

The procedure is summarized in Algorithm~\ref{algo:concreteness}. 
\begin{algorithm}[t]
\SetAlgoLined
\SetKwInOut{Input}{Output}
\KwIn{ list of normalized concreteness scores $\boldsymbol{a} = (a_1, a_2, \ldots, a_m)$, hyperparameter $\tau$}
\KwOut{Boundaries of constituents $B = \{(L_i, R_i)\}_{i=1, \ldots, m-1}$}
\For{$j=1$ to $m$} {
    $\textit{left}_j = j$ \\
    $\textit{right}_j = j$
}
 \While{$\textit{len}(\boldsymbol{a}) > 1$}{
  $p = \argmax_{j} a_{j} + \tau a_{j+1}$ \\
  add $(\textit{left}_p, \textit{right}_{p+1})$ to $B$ \\
  $\boldsymbol{a} = \boldsymbol{a}_{< p}$ + ($\frac{a_p + a_{p+1}}{2}$) + $\boldsymbol{a}_{>p+1}$ \\
  $\textbf{\textit{left}} = \textbf{\textit{left}}_{< p}$ + ($\textit{left}_p$) + $\textbf{\textit{left}}_{>p+1}$\\
  $\textbf{\textit{right}} = \textbf{\textit{right}}_{< p}$ + ($\textit{right}_{p+1}$) + $\textbf{\textit{right}}_{>p+1}$
 }
 \caption{\label{algo:concreteness} Constituency parsing based on concreteness estimation.}
\end{algorithm}
\begin{figure*}[!t]
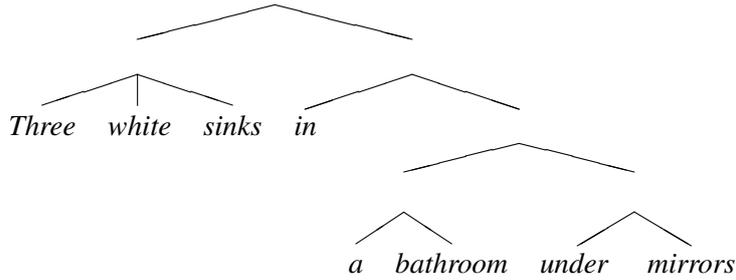
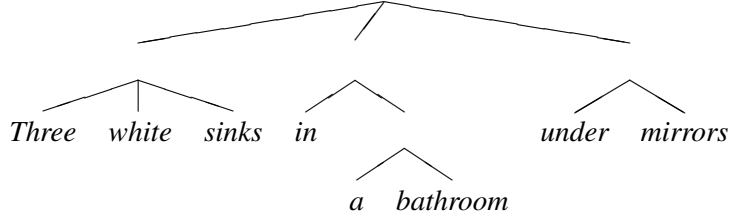

    \centering
    \begin{subfigure}[h]{0.6 \textwidth}
        \centering
        \begin{parsetree}
            (..
                (..
                    `Three'
                    `white' 
                    `sinks' 
                ) 
                (..
                    `in'
                    (..
                        (..
                            `a'
                            `bathroom'
                        ) 
                        (..
                            `under '
                            `mirrors' 
                        ) 
                    ) 
                ) 
            )
        \end{parsetree}
        \caption{Constituency parse tree labeled by Benepar \cite{kitaev2018constituency}. }
    \end{subfigure}
    \begin{subfigure}[h]{0.6 \textwidth}
        \centering
        \begin{parsetree}
            (..
                (..
                    `Three'
                    `white' 
                    `sinks' 
                ) 
                (..
                    `in'
                    (..
                        `a' 
                        `bathroom'
                    ) 
                ) 
                (.. 
                    `under' 
                    `mirrors' 
                ) 
            )
           
        \end{parsetree}
        \caption{Manually labeled constituency parse tree.}
    \end{subfigure}
    
    \caption{A failure example by Benepar, where it fails to parse the noun phrase ``three white sinks in a bathroom under mirrors'' -- according to human commonsense, it is much more common for sinks, rather than a bathroom, to be under mirrors. However, most of the constituents (e.g., ``\textit{three white sinks}'' and ``\textit{under mirrors}'') are still successfully extracted by Benepar. }
    \label{fig:ambiguity}
\end{figure*}

\section{Details of Manual Ground Truth Evaluation}
\label{appendix:evaluation}
It is important to confirm that the constituency parse trees of the MSCOCO captions produced by Benepar \cite{kitaev2018constituency} are of high enough qualities, so that they can serve as reliable ground truth for further evaluation of other models. 
To verify this, we randomly sample 50 captions from the MSCOCO test split, and manually label the constituency parse trees without reference to either Benepar or the paired images, following the principles by \citet{bies1995bracketing} as much as possible.\footnote{~The manually labeled constituency parse trees are publicly available at \url{https://ttic.uchicago.edu/~freda/vgnsl/manually_labeled_trees.txt}}
Note that we only label the tree structures without constituency labels (e.g., {\tt NP} and {\tt PP}). Most failure cases by Benepar are related to human commonsense in resolving parsing ambiguities, e.g., prepositional phrase attachments (Figure~\ref{fig:ambiguity}). 

We compare the manually labeled trees and those produced by Benepar \cite{kitaev2018constituency}, and find that the $\text{F}_1$ score between them are 95.65. 

\section{Concreteness by Different Models}
\label{appendix: concreteness}
\subsection{Correlation between Different Concreteness Estimations}
We report the correlation of different methods for concreteness estimation, shown in (\tbl{table:concreteness-others}). The concreteness given by \citet{turney2011literal} and \citet{brysbaert2014concreteness} highly correlate with each other. The concreteness scores estimated on multi-modal dataset \citep{hessel2018quantifying} also moderately correlates with the aforementioned two methods \citep{turney2011literal,brysbaert2014concreteness}. Compared to the concreteness estimated by \citet{hessel2018quantifying}, the one estimated by our model has a stronger correlation with the scores estimated from linguistic data \cite{turney2011literal,brysbaert2014concreteness}.

\subsection{Concreteness Scores of Sample Words by Different Methods}
We present the concreteness scores estimated or labeled by different methods in Figure~\ref{figure:concreteness-words}, which qualitatively shows that different methods correlate with others well.

\section{Sample Trees Generated by \model}
\label{appendix: example-trees}
\fig{figure:example-trees} shows the sample trees generated by \model with the head-initial inductive bias (\modelhi). All captions are chosen from the MSCOCO test set.
\begin{figure*}[t]
    \centering
        \begin{subfigure}[t]{\textwidth}
            \centering
            \begin{parsetree}( .. ( .. `a' `kitchen' ) ( .. `with' ( .. ( .. `two' `windows' ) ( .. `and' ( .. `two' ( .. `metal' `sinks' ) ) ) ) ) )\end{parsetree}
            \caption{a kitchen with two windows and two metal sinks}
        \end{subfigure}

        \begin{subfigure}[t]{\textwidth}
            \centering
            \begin{parsetree}( .. ( .. `a' ( .. `blue' ( .. `small' `plane' ) ) ) ( .. `standing' ( .. `at' ( .. `the' `airstrip' ) ) ) )\end{parsetree}
            \caption{a blue small plane standing at the airstrip}
        \end{subfigure}

        \begin{subfigure}[t]{\textwidth}
            \centering
            \begin{parsetree}( .. ( .. ( .. `young' `boy' ) `sitting' ) ( .. `on' ( .. `top' ( .. `of' ( .. `a' `briefcase' ) ) ) ) )\end{parsetree}
            \caption{young boy sitting on top of a briefcase}
        \end{subfigure}

        \begin{subfigure}[t]{\textwidth}
            \centering
            \begin{parsetree}( .. ( .. ( .. `a' ( .. `small' `dog' ) ) `eating' ) ( .. ( .. `a' `plate' ) ( .. `of' `broccoli' ) ) )\end{parsetree}
            \caption{a small dog eating a plate of broccoli}
        \end{subfigure}
    
\end{figure*}
\addtocounter{figure}{-1}
\begin{figure*}[t]
        \begin{subfigure}[t]{\textwidth}
        \addtocounter{subfigure}{4}
            \centering
            \begin{parsetree}( .. ( .. ( .. ( .. `a' `building' ) ( .. `with' ( .. `a' ( .. `bunch' ( .. `of' `people' ) ) ) ) ) ( .. `standing' `around' ) ) `it' )\end{parsetree}
            \caption{a building with a bunch of people standing around it}
        \end{subfigure}

        \begin{subfigure}[t]{\textwidth}
            \centering
            \begin{parsetree}( .. ( .. ( .. `a' `horse' ) `walking' ) ( .. `by' ( .. ( .. `a' `tree' ) ( .. `in' ( .. `the' `woods' ) ) ) ) )\end{parsetree}
            \caption{a horse walking by a tree in the woods}
        \end{subfigure}

        \begin{subfigure}[t]{\textwidth}
            \centering
            \begin{parsetree}( .. ( .. ( .. `the' ( .. `golden' `waffle' ) ) ( .. `has' ( .. `a' `banana' ) ) ) ( .. `in' `it' ) )\end{parsetree}
            \caption{the golden waffle has a banana in it .}
        \end{subfigure}

        \begin{subfigure}[t]{\textwidth}
            \centering
            \begin{parsetree}( .. ( .. ( .. `a' `bowl' ) ( .. `full' ( .. `of' `oranges' ) ) ) ( .. `that' ( .. `still' ( .. `have' `stems' ) ) ) )\end{parsetree}
            \caption{a bowl full of oranges that still have stems}
        \end{subfigure}

\end{figure*}
\addtocounter{figure}{-1}
\begin{figure*}[t]
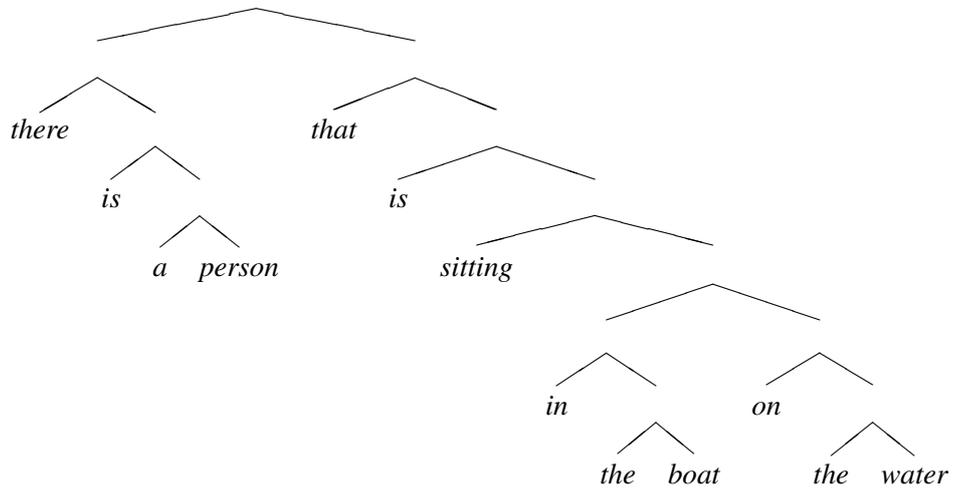


        \begin{subfigure}[t]{\textwidth}
        \addtocounter{subfigure}{8}
            \centering
            \begin{parsetree}( .. ( .. `there' ( .. `is' ( .. `a' `person' ) ) ) ( .. `that' ( .. `is' ( .. `sitting' ( .. ( .. `in' ( .. `the' `boat' ) ) ( .. `on' ( .. `the' `water' ) ) ) ) ) ) )\end{parsetree}
            \caption{there is a person that is sitting in the boat on the water}
        \end{subfigure}

        \begin{subfigure}[t]{\textwidth}
            \centering
            \begin{parsetree}( .. ( .. ( .. `a' `sandwich' ) ( .. `and' `soup' ) ) ( .. `sit' ( .. `on' ( .. `a' `table' ) ) ) )\end{parsetree}
            \caption{a sandwich and soup sit on a table}
        \end{subfigure}

        \begin{subfigure}[t]{\textwidth}
            \centering
            \begin{parsetree}( .. ( .. `a' ( .. `big' `umbrella' ) ) ( .. `sitting' ( .. `on' ( .. `the' `beach' ) ) ) )\end{parsetree}
            \caption{a big umbrella sitting on the beach}
        \end{subfigure}

    \caption{Examples of parsing trees generated by \model.}
    \label{figure:example-trees}
\end{figure*}

\end{document}